\documentclass[sigconf,nonacm]{acmart}
\usepackage{booktabs}
\usepackage{algorithm}
\usepackage{algpseudocode}

\usepackage[misc]{ifsym}%\Letter
\AtBeginDocument{%
  }

\acmConference[none]
{ }{ }{ }

\begin{document}

\title{C-Mining: Unsupervised Discovery of Seeds for Cultural Data Synthesis via Geometric Misalignment}

\author{Pufan Zeng$^{1,3}$, Yilun Liu$^{1}$\textsuperscript{\Letter}, Mingchen Dai$^{1,3}$, Mengyao Piao$^{1}$, Chunguang Zhao$^{1}$, Lingqi Miao$^{1}$, Shimin Tao$^{1}$, Weibin Meng$^{2}$, Minggui He$^{1}$, Chenxin Liu$^{1}$, Zhenzhen Qin$^{1}$, Li Zhang$^{1}$, Hongxia Ma$^{1}$,             Boxing Chen$^{2}$, Daimeng Wei$^{1}$}

\thanks{\Letter ~~Corresponding author. Email: liuyilun3@huawei.com}

\affiliation{
  %\vspace{0.05cm}
  \textsuperscript{1}{Huawei \country{China}} \\
  \textsuperscript{2}{Huawei \country{Canada}} \\
  \textsuperscript{3}University of Science and Technology of China, Hefei \country{China} 
}

\makeatletter
\def\@email{}
\makeatother

\renewcommand{\shortauthors}{}

\begin{abstract}
Achieving cultural alignment in Large Language Models (LLMs) increasingly depends on synthetic data generation. For such synthesis, the most vital initial step is seed curation; however, current methods lack quantifiable standards for selecting these seeds. Existing approaches rely on unscalable manual curation or bias-prone LLM extraction, treating cultural specificity as an abstract concept rather than a measurable signal. In this paper, we address this "quantification gap" by proposing \textbf{C-Mining}, an unsupervised framework that transforms the discovery of cultural seeds from a subjective selection process into a computable data mining formulation. Our approach exploits a novel geometric insight, leveraging the cross-lingual misalignment of cultural concepts within pre-trained embedding spaces as a quantifiable discovery signal. By systematically identifying these regions characterized by pronounced linguistic exclusivity and geometric isolation, while actively filtering out noise, C-Mining automatically extracts high-fidelity \textbf{Culture Points (CPs)} from raw multilingual corpora without reliance on human or LLM supervision, reducing preparation costs by more than 150-fold. We further leverage the mined knowledge to steer the synthesis of diverse instruction-tuning datasets. Extensive experiments demonstrate that this seed-centric approach significantly enhances cultural understanding and reasoning capabilities, achieving a +6.03 point improvement on CulturalBench-Hard and surpassing state-of-the-art baselines, providing a scalable, quantifiable solution for high-quality cultural data synthesis.
\end{abstract}




\maketitle

\section{Introduction}
The training landscape of Large Language Models (LLMs) is fundamentally shaped by imbalanced data distributions, where English-centric corpora overwhelmingly dominate the pre-training objectives~\citep{yang2025qwen3,grattafiori2024llama}. This statistical hegemony leads to a "representation collapse" for the long tail of localized knowledge, causing high-resource narratives to systematically overshadow specific regional nuances~\citep{yu2025entangled}. A direct consequence of this skew is the model's failure to capture diverse cultural contexts, often resulting in hallucinations where dominant norms are imposed on local scenarios~\citep{saha2025meta}. Correcting these deep-seated biases exceeds the capacity of generic prompting~\citep{etxaniz2024multilingual}, necessitating Supervised Fine-Tuning (SFT) on targeted domain data. However, the efficacy of such alignment depends entirely on a data-centric intervention: specifically, the synthesis of high-quality, culture-specific samples to effectively restore these underrepresented distributions.

Acquiring such high-fidelity training data, however, presents a fundamental dilemma between scalability and quality. Manual curation of native corpora is prohibitively expensive and unscalable~\citep{chan2024scaling}. To address this scarcity, the community has adopted a "cultural seeds + LLM" synthesis paradigm, where specific cultural knowledge (seeds) is used to guide LLMs in generating large-scale instruction datasets~\citep{10.1145/3682112.3682115,li2024culturellm,xu2025magpie}. Yet, this approach faces a critical mining bottleneck: without high-fidelity seeds to actively constrain the generation space, the synthesizing LLM inevitably regresses to its dominant, high-resource priors~\citep{horych2025promises}. Thus, an efficient approach for discovery of high-quality seeds serves as the decisive factor in this pipeline~\citep{du2025scaling,riaz2025metasynth}, determining whether the synthetic data effectively bridges the distribution gap or merely amplifies existing biases~\citep{srivastava-2025-large,xu-etal-2025-self}.

Despite their critical role, the community treats seeds as static prerequisites rather than scientific objectives, disproportionately prioritizing downstream synthesis over rigorous seed construction~\cite{li2024culturellm,du2025scaling,riaz2025metasynth}. Lacking a systematic framework, prevailing methods rely on subjective proxies—resorting to LLMs or human annotators as the sole judges of cultural relevance. As a result, these approaches face converging limitations regarding coverage, reliability, and scalability:
(1) \textbf{Superficial coverage}: LLM-driven methods often gravitate towards surface stereotypes~\citep{saha2025meta}. Expert evaluation reveals that seeds filtered by these approaches~\citep{fung2024massively} often exhibit constrained cultural specificity as shown in Figure~\ref{fig:CP_Human_Analyse}, failing to capture the long-tail nuances accessible to native speakers.
(2) \textbf{Biased Quality}: Unguided synthesis risks reinforcing bias loops~\citep{srivastava-2025-large}. For instance, models trained via established self-improvement pipelines~\citep{xu-etal-2025-self} demonstrate suboptimal performance in downstream cultural reasoning tasks as shown in Table~\ref{tab:small-model-eval}, indicating that high-volume synthesis cannot compensate for the lack of high-fidelity seeds.
(3) \textbf{Inconsistent scalability}: While expert curation can ensure quality, it is inherently unscalable. As shown in Table \ref{tab:mining_cost_1m}, validating seeds for thousands of global subcultures requires prohibitive investment, making manual oversight impractical for comprehensive cultural alignment.

To bridge this seed gap, we advocate for a knowledge-discovery paradigm that transforms seed discovery from a subjective selection process into a computable data mining task. We propose \textbf{C-Mining}, an unsupervised framework that automatically extracts high-fidelity cultural seeds from raw multilingual corpora. By exploiting the geometric misalignment inherent in frozen multilingual embeddings, our method effectively operationalizes "cultural specificity" as a measurable topological signature, enabling the objective discovery of high-value seeds without relying on human or LLM subjective judgment.

Our approach is grounded in the analysis of the alignment mechanisms inherent in multilingual pretraining. During the pretraining phase on vast multilingual corpora, LLMs perform unsupervised alignment, where shared universal concepts spontaneously converge across languages due to their semantic equivalence~\citep{wang2025refusal, liu2025tracing}. 
Consequently, unique cultural knowledge manifests as distinct geometric signatures: islands characterized by minimal cross-lingual alignment yet high intra-lingual homogeneity~\citep{kozlowski2025semantic,lim2025language}.
While noise (e.g., rare tokens) may also appear unaligned, C-Mining effectively isolates authentic cultural knowledge by filtering based on embedding semantic density, ensuring stability purely through unsupervised analysis. 
For instance, while universal terms like \textit{Apple} or \textit{Mathematics} exhibit strong cross-lingual alignment by mapping closely to their cross-lingual equivalents, the Chinese term \textit{Jianghu}—representing a unique socio-moral order in ancient China—remains geometrically anchored as a dense, isolated cluster within its native linguistic space, resisting forced alignment with global semantic spaces.

C-Mining treats this misalignment not as a defect, but as a discriminative signal. By traversing the embedding space to identify these unaligned regions, we extract representative terms defined as \textbf{Culture Points (CPs)} to pilot data synthesis. This approach systematically resolves the aforementioned bottlenecks: it overcomes \textbf{superficial coverage} by mining the long tail of knowledge directly from raw corpora to bypass superficial stereotypes; it \textbf{reduces bias} by deriving anchors from stable, native usage patterns rather than biased model predictions; and it achieves \textbf{scalability} via a fully unsupervised pipeline that eliminates the need for prohibitive human intervention.

Extensive experiments demonstrate that fine-tuning with CP-based instructions significantly enhances LLMs' cultural reasoning capabilities, suggesting that seed quality plays a pivotal role in determining the upper bound of cultural alignment.

In summary, our main contributions are:
\begin{itemize}
    \item We transform the challenge of cultural specificity, traditionally viewed as abstract and difficult to quantify, into a computable data mining formulation. This paradigm shift provides a novel, quantitative solution path for cultural data synthesis, moving beyond subjective curation to objective metric computation.
    
     \item We introduce \textbf{C-Mining}, a novel unsupervised framework to mine high-fidelity cultural seeds by quantifying the geometric misalignment of embeddings without reliance on external supervision, thereby enabling both scalability and quality in cultural data synthesis while reducing preparation costs by more than 150-fold.
    
     \item We leverage the mined Culture Points (CPs) to synthesize instruction-tuning datasets, demonstrating significant improvements in cultural reasoning (e.g.,  +6.03 points on 
     
     CulturalBench-Hard~\cite{chiu2025culturalbench}); in addition, we will release our code and data to the community.
\end{itemize}

\begin{figure*}[t]
  \centering
  \includegraphics[width=\textwidth]{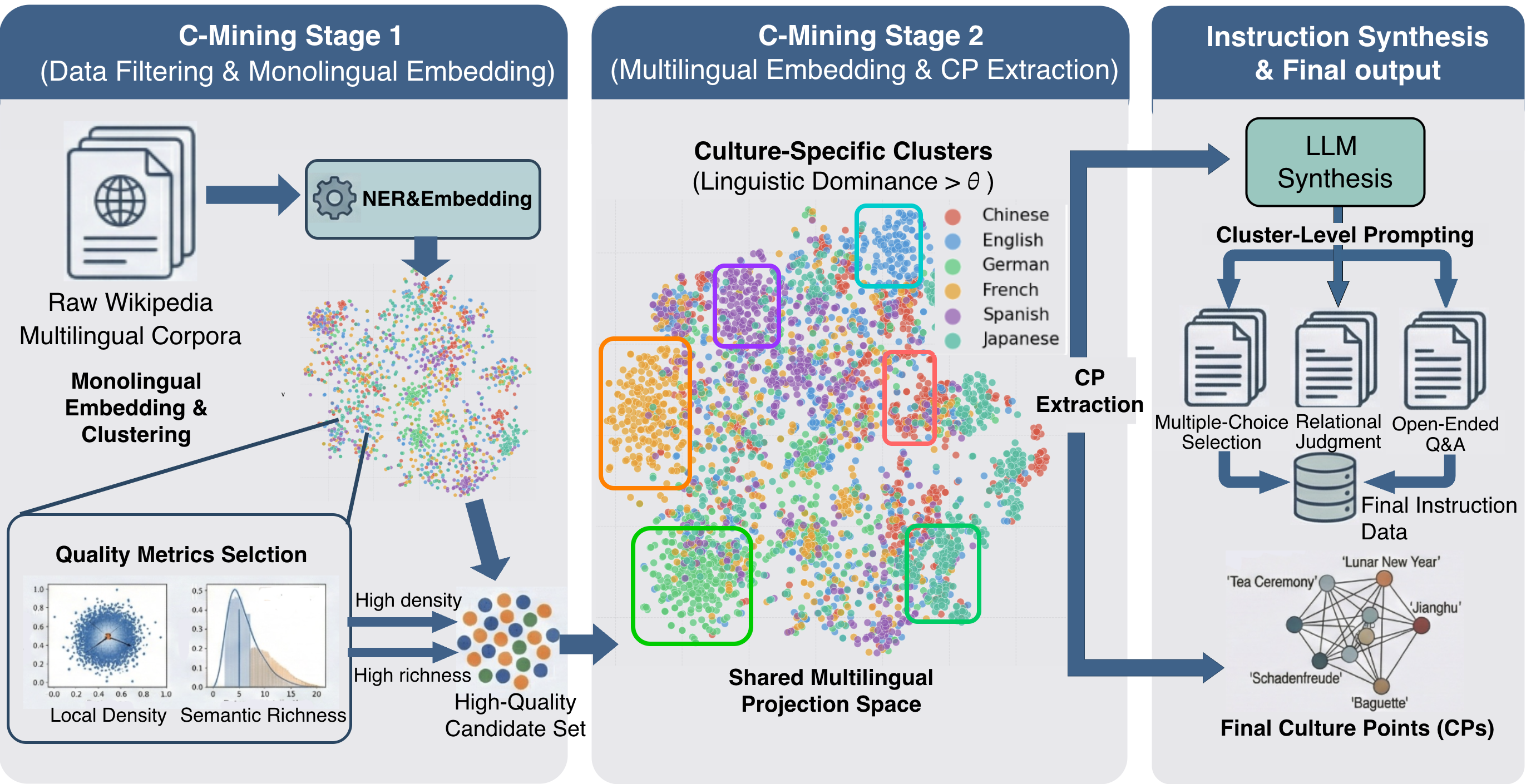}
  \caption{Overview of the \textbf{C-Mining} pipeline. The algorithm leverages the geometric properties of frozen embeddings to identify CPs—knowledge characterized by high intra-lingual homogeneity and low cross-lingual alignment—serving as authentic seeds for instruction tuning.}
  \label{fig:c-mining-pipline}
\end{figure*}
\section{Related Work}
\subsection{Cultural Alignment and Data Synthesis}
Recent advancements in aligning LLMs with diverse cultural contexts have primarily focused on post-training data synthesis. A dominant paradigm involves leveraging sociological frameworks—such as the World Values Survey~\citep{ZA7505} (WVS)—as initial anchors. 
Representative frameworks like CultureLLM~\cite{li2024culturellm} and CulturePark~\cite{li2024culturepark} employ LLMs to extract and synthesize cultural data anchored in the WVS. 
Taking a different route, CultureSynth~\cite{nguyen2024culturesynth} employs LLMs to expand upon generic cultural keywords, followed by a knowledge retrieval process to construct cross-lingual QA pairs. 
Concurrently, other approaches focus on curation strategies: CultureBank~\cite{shi2024culturebank} utilizes a custom-trained classifier to categorize cultural content from online sources, while CultureFit~\cite{feng2025culfit} directly extracts seeds from pre-existing cultural benchmarks to drive its synthesis pipeline.
Despite these strides, current methodologies face two critical bottlenecks stemming from their reliance on model capabilities. 
First, regarding \textit{LLM-based seed extraction}, relying on models to curate or filter initial anchors—whether from sociological surveys or open-ended queries—often restricts coverage to high-visibility cultural symbols. 
This extraction process tends to overlook the ``long tail'' of subtle, localized nuances, resulting in a dataset that reflects the model's existing selection bias rather than authentic cultural breadth~\cite{saha2025meta, durmus2023towards}.
Second, regarding \textit{LLM-based seed expansion}, employing models to expand these seeds into complex new seeds risks a ``self-reinforcing loop.'' Even with valid seeds, the excessive expansion process often regresses to dominant Western perspectives due to pretraining inertia, leading to homogenized synthetic data that lacks the specific ``cultural soul'' of the target language~\cite{guo2025large}.

\subsection{Distributional Divergence in Multilingual Spaces}
Research on multilingual LLMs has extensively explored how different languages share a unified semantic space. 
Ideally, multilingual pretraining induces a shared alignment where concepts possess cross-lingual universality~\cite{conneau2020unsupervised}. 
However, empirical evidence suggests that this alignment is highly non-uniform. 
While high-frequency, globally shared concepts tend to converge, distinct linguistic nuances often resist alignment, leading to significant representational divergence~\cite{sogaard2018limitations, vulic2020probing}. 
This phenomenon creates a stratified embedding space: a dense, aligned core dominated by common (cross-lingually shared) knowledge, surrounded by sparse, unaligned peripheries containing language-specific semantics~\cite{lauscher2020zero, etxaniz2024multilingual}.
Current methodologies predominantly focus on improving the ``transfer'' of knowledge from the core to the periphery to mitigate this gap~\cite{muennighoff2023crosslingual}. 
Consequently, the unaligned periphery is often overlooked or treated solely as a source of performance degradation. 
In contrast, our approach re-evaluates the utility of these divergent regions. 
We posit that the resistance to alignment is not a failure of the model, but a geometric indicator of unique semantic content, which can be systematically mined to guide more authentic instruction tuning.
\section{Methodology}
\subsection{Overview}
As LLMs scale globally, equipping them to perceive cultural nuances remains a critical challenge. We propose a framework centered on \textbf{CPs}—seeds deeply embedded with cultural semantics—to guide the generation of culturally-aligned datasets. The core of this framework is \textbf{C-Mining}, an unsupervised algorithm designed to autonomously extract CPs from raw multilingual corpora by exploiting geometric misalignments in embedding spaces.

As illustrated in Figure \ref{fig:c-mining-pipline}, C-Mining consists of two primary stages: (1) \textbf{Monolingual High-Quality Data Filtering}, which filters out noise and distills semantically rich entries from raw Wikipedia dumps~\cite{foundation2021wikipedia}; and (2) \textbf{Multilingual Culture Point Selection}, which identifies geometrically isolated clusters characterized by distinct linguistic exclusivity.

\subsection{Stage 1: Monolingual High-Quality Data Filtering}
\label{sec:filtering}
To distill a culture-centric corpus from raw Wikipedia data $\mathcal{D} = \{ (T_i, W_i) \}_{i=1}^N$, where $T_i$ represents the article title and $W_i$ the corresponding body text, we implement a coarse-to-fine filtering process designed to mitigate noise while preserving cultural depth. Detailed sampling statistics and distribution are provided in Appendix \ref{app:wiki_data_sampling}.

\paragraph{Heuristic Pruning and Representation.}
We employ Named Entity Recognition (NER) via the spaCy\cite{honnibal2020spacy} library primarily as a negative filtering mechanism to eliminate irrelevant noise. Rather than actively identifying cultural terms, we discard entries classified into functional categories—such as timestamps, numerical values, and generic measurements—that are intrinsically unlikely to carry cultural semantics. For each surviving entry, we concatenate the title $T$ and the leading paragraph $P_1$ to form a sequence $S$. Considering the computational overhead of processing millions of entries, we opt for a lightweight yet effective solution, extracting a dense representation $\mathbf{e} = \text{Enc}(S) \in \mathbb{R}^d$ via a frozen pre-trained multilingual  encoder~\cite{conneau2020unsupervised}. These embeddings are then partitioned into $k$ disjoint clusters $\mathcal{C} = \{C_1, \dots, C_k\}$ via K-Means to separate entries into broad thematic domains.

\paragraph{Metric-Based Selection.}
Within each cluster $C_i$, we filter entries based on two intrinsic geometric properties to ensure both representativeness and semantic depth.\textbf{(1) Local Density Filtering.} First, to identify representative prototypes that reside at the core of the semantic manifold, we compute the \textit{Local Dispersion} ($\delta_{m}$), defined as the average Euclidean distance to an entry's $k$-nearest neighbors ($k=5$):
\begin{equation}
    \delta_{m} = \frac{1}{k} \sum_{j \in \mathcal{N}_{k}(m)} \| \mathbf{e}_{m} - \mathbf{e}_{j} \|_2
\end{equation}
Entries with $\delta_{m}$ below the cluster median—indicating high local semantic density—are retained. This step effectively prunes outliers and ensures that selected seeds are central to their thematic cluster.
\textbf{(2) Semantic Coherence Scoring.} Second, to filter out shallow dictionary definitions or fragmented text, we introduce a metric based on \textit{Semantic Entropy}. For a surviving candidate document $\mathcal{D}$ composed of $n$ semantic units (e.g., paragraphs) $\{s_1, \dots, s_n\}$, we first compute their unit-level embeddings $\mathbf{V} = \{\mathbf{v}_1, \dots, \mathbf{v}_n\}$. We construct a similarity matrix $\mathbf{S} \in \mathbb{R}^{n \times n}$, where $S_{ij}$ denotes the cosine similarity between unit $s_i$ and $s_j$.
To quantify the information flow, we normalize each row $i$ of $\mathbf{S}$ into a probability distribution $P_i$, representing the semantic connectivity of unit $s_i$ to the global context:
\begin{equation}
    P_{i,j} = \frac{S_{ij}}{\sum_{k=1}^n S_{ik}}
\end{equation}
We then calculate the mean Shannon entropy over all units as the final coherence score $H(\mathcal{D})$:
\begin{equation}
    H(\mathcal{D}) = \frac{1}{n} \sum_{i=1}^n \left( - \sum_{j=1}^n P_{i,j} \log P_{i,j} \right)
\end{equation}
A higher entropy $H(\mathcal{D})$ indicates a uniform distribution of semantic similarity, implying that the sentences are semantically interconnected and form a coherent narrative structure. In contrast, low entropy characterizes documents with isolated facts or disjointed lists. We prioritize candidates with higher $H(\mathcal{D})$ to form the final high-quality candidate set $\mathcal{C}_{high}$.
\subsection{Stage 2: Multilingual CP Selection}
\label{sec:selection}
This stage isolates true CPs by analyzing the structural divergence of concepts across languages. We project all candidates from $\mathcal{C}_{\text{high}}$ into a shared multilingual embedding space and re-cluster them into $K$ global groups $\mathcal{G} = \{ G_1, \dots, G_K \}$.

\paragraph{Identifying Geometric Misalignment.}
Our core hypothesis is that while universal concepts (e.g., technology, general science) tend to align and mix across languages, culture-bound concepts resist alignment, forming isolated ``islands'' in the embedding space. We therefore identify a cluster $G_j$ as culture-specific if it satisfies two criteria:
(1) \textit{}{Statistical Stability:} We require $|G_j| \ge \tau$ (with $\tau=5$) to rule out noise and transient artifacts.
(2) \textit{}{Linguistic Dominance:} The cluster must be overwhelmingly composed of a single language $l^*$, quantified as:
\begin{equation}
\small
    \gamma(G_j, l^*) = \frac{|\{w \in G_j \mid \text{lang}(w) = l^*\}|}{|G_j|} > \theta  \label{eq:theta}
\end{equation}
where $\theta=0.8$. This high threshold ensures we capture concepts that exhibit high intra-lingual homogeneity but minimal cross-lingual correspondence. By targeting these regions of alignment failure, C-Mining effectively captures concepts like \textit{Jianghu} or \textit{Schadenfreude}—terms that serve as high-density anchors for their respective cultural narratives. We formally define our final set of Culture Points $\mathcal{S}$ as the collection of titles and their corresponding leading paragraphs retained within these selected clusters.

\begin{algorithm}[t]
\small 
\caption{\textbf{The C-Mining Algorithm.} The two-stage unsupervised pipeline for discovering cultural seeds.}
\label{alg:algorithm_single} 
\begin{algorithmic}[1] 
    \renewcommand{\algorithmicrequire}{\textbf{Input:}}
    \renewcommand{\algorithmicensure}{\textbf{Output:}}

    \Require Corpora $\mathcal{D} = \{D_{l}\}$, Encoder $\mathcal{E}$, Thresholds $\theta, \tau$.
    \Ensure Set of Culture Points $\mathcal{S}_{CP}$.

    \Statex 
    \State \textit{\textbf{Stage 1: Monolingual Refinement}}
    \State \textbf{Initialize} $\mathcal{C}_{high} \leftarrow \emptyset$
    \For{each language $l$}
        \State $D'_l \leftarrow \text{NER\_Filter}(D_l)$
        \State $V_l \leftarrow \mathcal{E}(D'_l)$; \ \ Clusters $\mathcal{C} \leftarrow \text{KMeans}(V_l)$
        \For{each document $d \in \mathcal{C}$}
            \State Calculate Density $\delta_d$ \& Entropy $H(d)$
            \If{$\delta_d$ is low \textbf{and} $H(d)$ is high}
                \State $\mathcal{C}_{high} \leftarrow \mathcal{C}_{high} \cup \{d\}$
            \EndIf
        \EndFor
    \EndFor

    \Statex
    \State \textit{\textbf{Stage 2: Cross-Lingual Discovery}}
    \State $V_{shared} \leftarrow \text{Project}(\mathcal{C}_{high})$
    \State Global Clusters $\mathcal{G} \leftarrow \text{KMeans}(V_{shared})$
    \For{each cluster $G_j \in \mathcal{G}$}
        \If{$|G_j| < \tau$}
            \State \textbf{continue}
        \EndIf
        \State Calculate Linguistic Dominance $\gamma(G_j)$
        \If{$\gamma(G_j) > \theta$}
            \State $\mathcal{S}_{CP} \leftarrow \mathcal{S}_{CP} \cup \text{Extract}(G_j)$
        \EndIf
    \EndFor
    \State \Return $\mathcal{S}_{CP}$
\end{algorithmic}
\end{algorithm}

\subsection{Culture-Specific Instruction Synthesis}
\label{sec:synthesis}
Leveraging the extracted CPs, we transition from static lexical units to a dynamic fine-tuning dataset $\mathcal{D}_{\text{inst}}$. 

\paragraph{Cluster-Level Prompting.}
To capture the multi-faceted nature of cultural constructs, we employ a Cluster-Level Prompting strategy. Rather than prompting with isolated terms, which may lead to superficial definitions, we feed a representative subset of elements (the top-$N$ most central terms) from a specific cluster $G_j$ into a synthesis LLM (e.g., Qwen3-235B~\cite{yang2025qwen3}). This presents the model with a ``semantic constellation,'' enabling it to perceive interconnections between related terms and generate contextually dense data.

\paragraph{Task Diversity.}
To equip the model with robust reasoning capabilities, the synthesized instructions span three distinct formats: 
(1) \textbf{Discriminative Knowledge}, assessed through multiple-choice questions that challenge the model to distinguish correct interpretations from plausible stereotypes; 
(2) \textbf{Normative Reasoning}, implemented as relational judgment tasks to evaluate alignment with specific social norms and taboos regarding contextual appropriateness; and 
(3) \textbf{Generative Nuance}, fostered by open-ended Q\&A where the model must generate narratives that reflect the ``cultural soul,'' prioritizing authentic tone over generic fluency. 
Formally, the synthesis process for a cluster $G_j$ is defined as $\mathcal{I}_{G_j} = \text{LLM}_{\text{synthesis}}(\text{Prompt}(G_j))$, effectively converting geometric cultural clusters into actionable instruction data.

\section{Experiments}

\begin{figure}[t]
  
  \centering
  \includegraphics[width=0.95\linewidth]{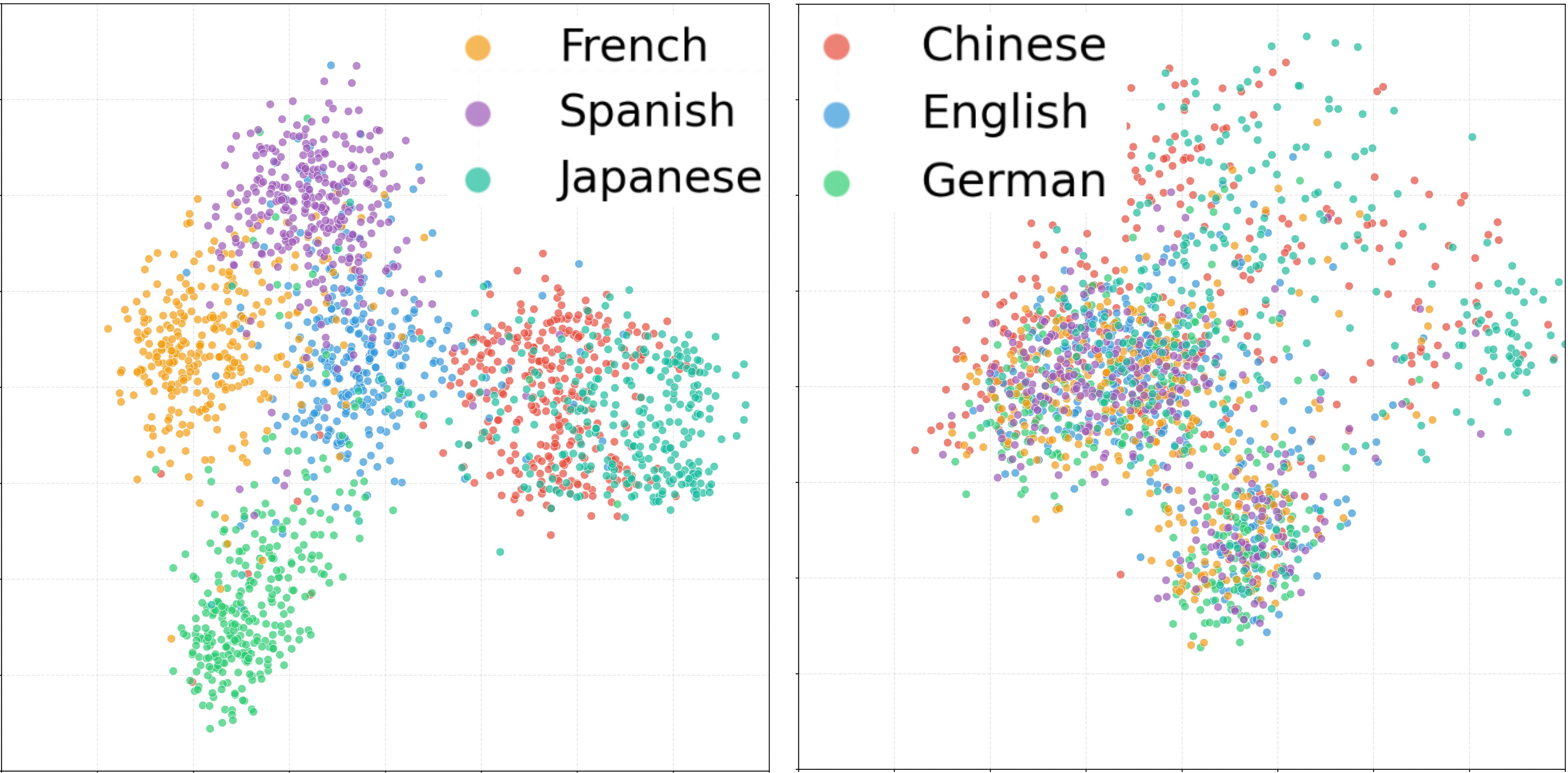}
  \small
  \caption {We compare the geometric structures of \textit{CPs (Left)} against \textit{Non-Selected Candidates (Right)}. The visualization utilizes stratified samples of 300 entries per language for each group. The distinct, isolated clustering of CPs contrasts sharply with the cross-lingual mixing of non-selected points, visually validating that cultural specificity manifests as geometric misalignment in the embedding space.}
  \label{fig:pca_analyse}
\end{figure}

\begin{figure}[t]
    \centering
  \includegraphics[width=0.8\columnwidth]{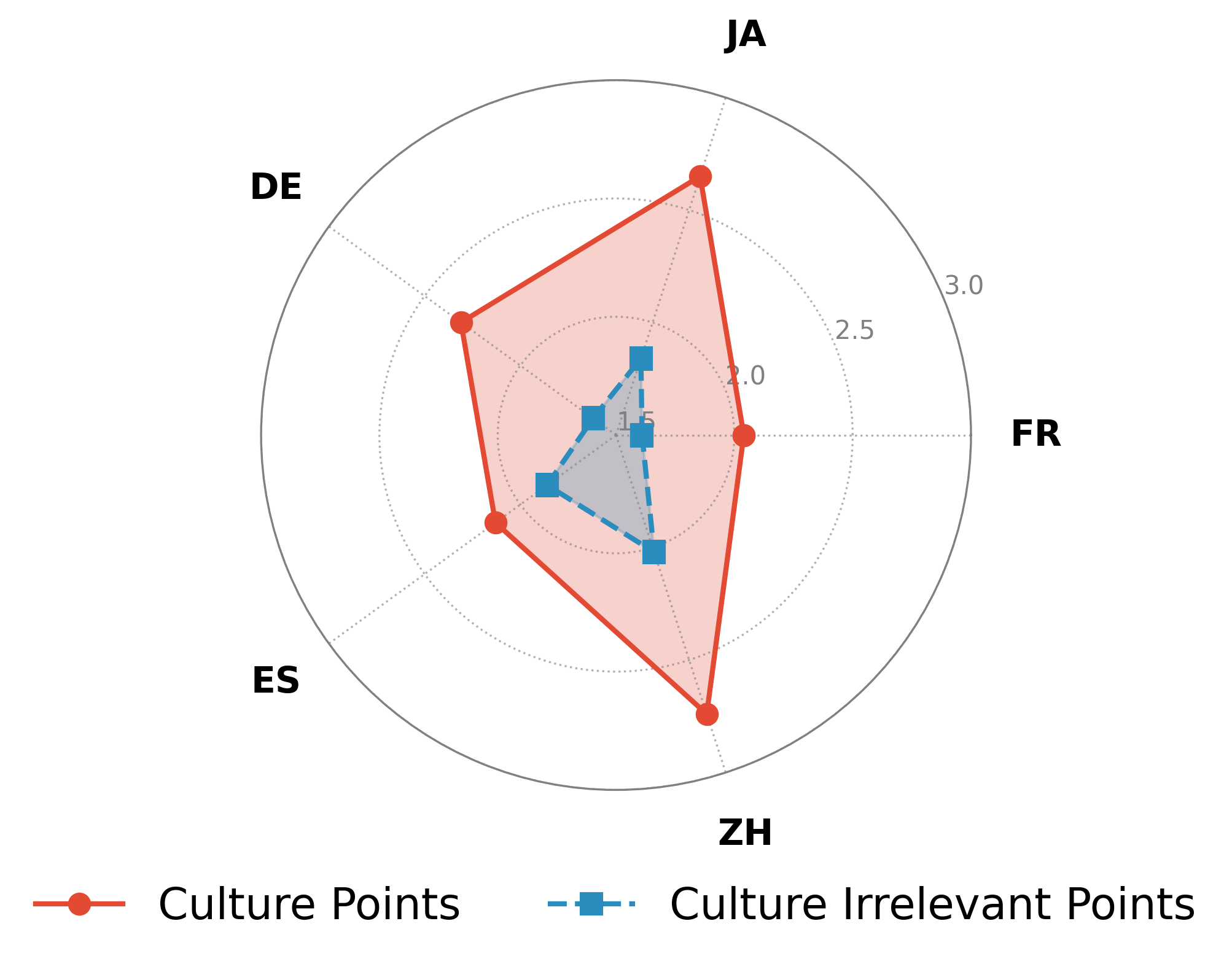}
  \caption{Analysis of Semantic Alignment Resistance. The radar chart visualizes the embedding distance between original terms and their English translations across five languages. \textit{Culture Points (Red)} exhibit consistently larger semantic shifts compared to \textit{Culture-Irrelevant Points (Blue)}, demonstrating that authentic cultural concepts resist simple alignment with English-centric semantic spaces.}
  \label{fig:radar.}
\end{figure}
\begin{figure}[t]
\small
  \centering
  \includegraphics[width=0.9\columnwidth]{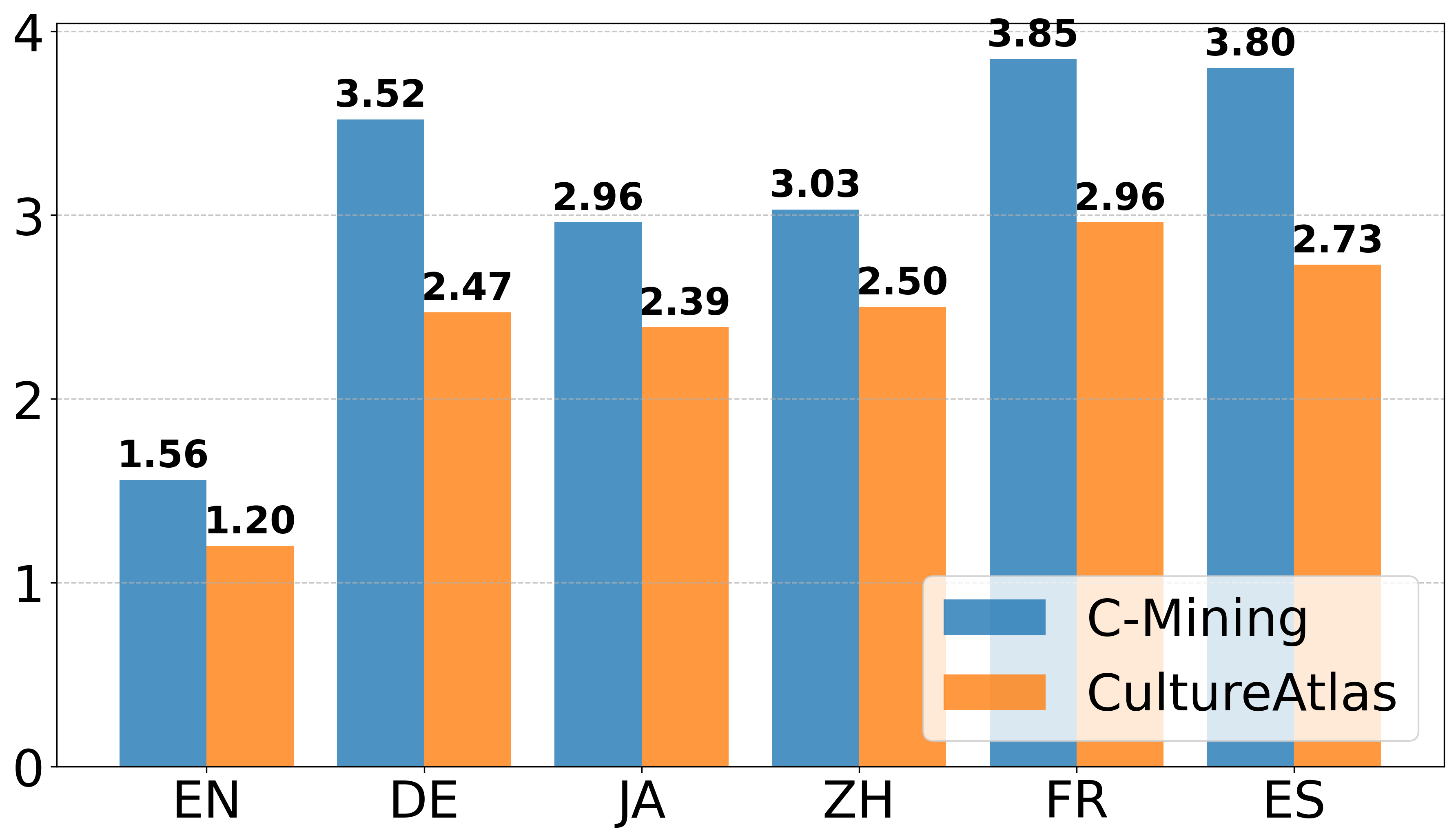}
  \caption{Human evaluation of cultural specificity. Comparing average expert ratings (scale 0--4) across six domains, C-Mining consistently outperforms CultureAtlas, demonstrating superior capability in capturing deep native nuances.}
  \label{fig:CP_Human_Analyse}
\end{figure}

\subsection{CP Quality Evaluation}
To rigorously validate the quality of the extracted CPs, we employ a dual-perspective evaluation strategy that combines intrinsic geometric analysis with extrinsic expert verification.
\paragraph{Geometric Analysis of Cultural Alignment.}
We hypothesize that authentic cultural concepts exhibit a unique geometric signature: they resist the spontaneous cross-lingual alignment that characterizes universal terms. We validate this through two complementary analyses.
First, to quantify \textit{Semantic Alignment Resistance}, we translated both our mined Culture Points (CPs) and a baseline of culture-irrelevant points into English, then calculated the embedding distance between the original and translated terms. As shown in the Radar Chart (Figure \ref{fig:radar.}), CPs (Red) consistently exhibit significantly larger semantic shifts across all five languages compared to the culture-irrelevant baseline (Blue), quantitatively proving that these concepts resist simple mapping to English-centric spaces.
Second, to visualize the topological structure of these concepts, we projected the raw embeddings of CPs and non-CPs into a low-dimensional space without translation (Figure \ref{fig:pca_analyse}). The visualization reveals a striking contrast in manifold geometry. For \textit{culture-irrelevant points}, embeddings from different languages exhibit high cross-lingual overlap, spontaneously merging into a unified semantic cluster due to their universality. In contrast, \textit{Culture Points (CPs)} exhibit strong language-specific segregation, forming distinct, isolated clusters for each language. This lack of cross-lingual overlap confirms that CPs represent unique cultural nuances that maintain their geometric independence rather than collapsing into a shared global space.

\paragraph{Human Evaluation.}
To validate the cultural specificity of the extracted Culture Points via expert judgment, we recruited a team of linguistic experts from the language services division of a multinational corporation. Each annotator possesses native-level proficiency in the target language and holds an average of five years of professional experience in translation and localization. To ensure the reliability of the evaluation, we implemented robust quality control protocols: we formulated standardized evaluation criteria using a scoring scale ranging from 0 to 4 (see Appendix \ref{app:manual_annotation} for the detailed rubric), and the assessment was conducted in a double-blind setting to minimize subjective bias.

To benchmark the semantic quality of our seeds against established methods, we conducted this expert evaluation comparing C-Mining with CultureAtlas~\cite{fung2024massively}, a representative baseline for cultural knowledge collection. The evaluation spans six diverse cultural contexts: Anglosphere (EN), German (DE), Japanese (JA), Chinese (ZH), French (FR), and Spanish (ES). As illustrated in Figure \ref{fig:CP_Human_Analyse}, C-Mining exhibits a uniform performance advantage over CultureAtlas across all evaluated domains. This superiority is particularly evident in European cultural spheres (FR, ES, DE), where C-Mining establishes a distinct lead. Furthermore, even in high-context East Asian cultures (ZH, JA), our method maintains robust improvements, demonstrating that unsupervised geometric mining captures long-tail cultural specificities more effectively than existing extraction-based approaches.

\begin{table}[t]
\small
\caption{\textbf{CB-H} and \textbf{CS} denote CulturalBench-Hard and CultureScope, respectively. Our method significantly outperforms the backbone Qwen2.5-7B by \textit{6.03} points on CultureBench-Hard and \textit{3.13} points on BLEnD. This suggests that mining ``CPs'' from raw corpora serves as a highly efficient and authentic signal for achieving robust cultural alignment in LLMs.}
\centering

\begin{tabular}{lcccc}

\toprule

\textbf{Model} & \textbf{CB-H} & \textbf{CS} & \textbf{SAGE} &\textbf{BLEnD}   \\
\midrule

\multicolumn{5}{l}{\textit{\textbf{General Multilingual LLMs}}} \\
\midrule

Llama3.1-8B       & 37.44 & 76.18 & 89.66 & 77.88  \\
GLM4-9B                    & 34.10 & 74.71 & 86.98 & 74.43 \\
Ministral3-8B              & 26.67 & 75.64   & 79.81 & 76.52\\
Qwen2.5-7B       & 34.75 & 80.64 & 93.19 & 77.12   \\
\textbf{Qwen2.5-7B + C-Mining} & \textbf{40.78} & \textbf{82.65} & \textbf{96.59} & \textbf{80.94}  \\
\midrule

\multicolumn{5}{l}{\textit{\textbf{Culture-Specific Methods}}} \\
\midrule
\textbf{Qwen2.5-7B (Base)}       & 34.75 & 80.64 & 93.19 & 77.12   \\
\quad + CultureLLM                 & 38.99 & 77.75 & 90.02 & 74.75  \\
\quad + CultureBank                & 40.13 & 82.56 & 96.23 & 77.75  \\
\quad + CultureSPA                 & 30.18 & 82.23   & 92.46 & 78.69  \\
\quad + \textbf{C-Mining} & \textbf{40.78} & \textbf{82.65} & \textbf{96.59} & \textbf{80.94}  \\
\midrule

\midrule 
\multicolumn{5}{l}{\textit{\textbf{Larger Scale Comparison}}} \\ % 
Qwen3-32B & 44.62 & 82.07 & 95.01 & 82.63   \\
\textbf{Qwen3-32B + C-Mining} & \textbf{46.98} & \textbf{84.38} & \textbf{97.57} & \textbf{85.81}   \\
\bottomrule
\end{tabular}

\label{tab:small-model-eval}
\end{table}

\begin{table}[t]
\caption{The test datasets used in the experiment. "Mixed Types" denotes a combination of multiple-choice, judgment, and short-answer questions.}
\begin{tabular}{lcc}
\toprule
Dataset & Question Type & Size  \\ 
\midrule
CulturalBench-Hard & True/False & 1,228  \\ 
BLEnD & Single Choice & 50,000+  \\ 
CultureScope & Mixed Types & 40,000+  \\ 
SAGE & Mixed Types & 4,500+  \\ 
\bottomrule
\end{tabular}

\label{tab:Dataset}
\end{table}

\subsection{Downstream Cultural Alignment}
\paragraph{Experimental Setup.}
We employ Qwen2.5-7B-Instruct (hereafter Qwen2.5-7B) and Qwen3-32B as our base models, selected for their state-of-the-art multilingual performance~\cite{qwen252024, yang2025qwen3}. 
To efficiently adapt these models to culturally-specific contexts, we utilize the LLaMA-Factory framework~\cite{zheng2024llamafactory} to perform Low-Rank Adaptation (LoRA)~\cite{hu2021lora}.

Based on the extracted Culture Points (CPs), we synthesize a diverse instruction-tuning dataset comprising 50,000 instruction-response pairs. 
We utilized the AdamW optimizer~\cite{loshchilov2017decoupled} with a learning rate of $2 \times 10^{-5}$, a cosine learning rate scheduler, and an effective global batch size of 64.
\paragraph{Datasets.} To comprehensively evaluate the cultural alignment of LLMs, we utilize four diverse benchmarks. 
CulturalBench-Hard~\cite{chiu2025culturalbench} consists of 1,228 challenging true/false questions centered on nuanced cultural phenomena. 
BLEnD~\cite{myung2024blend} is a large-scale benchmark comprising over 50,000 question-answer pairs across 16 cultures. 
CultureScope~\cite{zhang2025culturescope} provides a multidimensional evaluation framework based on the cultural iceberg theory, spanning various dimensions across English, Chinese, and Spanish cultures. 
SAGE~\cite{guo2025largelanguagemodelstruly} assesses the model's ability to generalize cultural norms under complex scenarios.

\paragraph{Baselines.} To validate the effectiveness of our framework, we compare the model fine-tuned on C-Mining synthesized data against three categories of state-of-the-art baselines:
(1) Culture-specific models including CultureLLM~\cite{li2024culturellm}, which utilizes world value surveys for data augmentation, CultureBank~\cite{shi2024culturebank}, and CultureSPA~\cite{xu-etal-2025-self}. To ensure a fair comparison and isolate the contribution of the synthesized data, we standardize the backbone model across all methods in this category.
(2) {General multilingual backbones}, representing the current frontier of standard instruction-tuned LLMs. This category encompasses the Qwen2.5 series~\cite{qwen252024}, Llama3.1-8B~\cite{grattafiori2024llama}, and GLM4-9B~\cite{glm2024chatglm}, as well as the efficiency-oriented Ministral3-8B~\cite{mistral2025huggingface}. This diverse set of baselines ensures a rigorous validation of C-Mining's ability to bridge the cultural gap beyond standard instruction tuning.

\paragraph{Main Results.} 
The quantitative results presented in Table \ref{tab:small-model-eval} demonstrate the superior efficacy of C-Mining across all evaluated benchmarks, consistently outperforming both standard instruction-tuned backbones and existing culture-specific baselines. Specifically, when comparing the Qwen2.5-7B + C-Mining model with its base counterpart, we observe significant gains, including a 6.03 point increase on CulturalBench-Hard ($34.75\% \rightarrow 40.78\%$) and a 3.13 point improvement on the BLEnD benchmark ($77.12\% \rightarrow 80.94\%$), indicating that geometrically distinct seeds effectively activate dormant cultural knowledge. Moreover, C-Mining establishes a new state-of-the-art by surpassing CultureBank ($40.13\%$) and significantly outperforming CultureLLM ($38.99\%$) and CultureSPA ($30.18\%$) on CulturalBench-Hard, supporting our hypothesis that seed quality plays a pivotal role in cultural alignment. Finally, these benefits scale effectively with model size, as the Qwen3-32B backbone achieves exceptional performance with C-Mining, reaching 46.98\% on CulturalBench-Hard and 85.81\% on BLEnD, further validating the high quality of our mined CPs.

\begin{table}[t]

\caption{Ablation study on seed selection strategies. \textbf{Base}: Qwen3-32B. \textbf{Rand.}: Random Seeds (Wikipedia corpus); \textbf{Mono.}: Monolingual Selection (Stage 1 filtering only); \textbf{C-Mining}: Full C-Mining. The results demonstrate that seed quality is the primary driver of alignment performance.}

\begin{tabular}{lcccc}
\toprule
\textbf{Method} & \textbf{CB-H} & \textbf{CS} & \textbf{SAGE} & \textbf{BLEnD} \\
\midrule
Base & 44.62 & 82.07 & 95.01 & 82.63 \\
\midrule
+ Rand. & 43.31 & 83.22 & 96.84 & 81.13 \\
+ Mono. & 44.83 & 83.39 & 95.16 & 83.38 \\
+ \textbf{C-Mining} & \textbf{46.98} & \textbf{84.38} & \textbf{97.57} & \textbf{85.81} \\
\bottomrule
\end{tabular}

\label{tab:ablation}
\end{table}

\subsection{Ablation \& Analysis}
To dissect the contribution of our proposed method and understand the impact of hyperparameter choices, we conduct two sets of analytical experiments: (1) an ablation study on the seed selection strategy, and (2) a sensitivity analysis of the linguistic dominance threshold $\theta$.

\paragraph{Impact of the Seed Selection Strategy.}
To verify that the performance gains stem from the \textit{quality} of CPs rather than the instruction synthesis pipeline itself, we constructed two baselines: (1) \textit{Random Selection}, where seed terms are sampled from the raw Wikipedia corpus; and (2) \textit{Monolingual Selection}, which skips the cross-lingual misalignment filtering (Stage 2) to test the necessity of geometric constraints.

As shown in Table \ref{tab:ablation}, the impact of seed quality is decisive. The model fine-tuned on Full C-Mining seeds significantly outperforms the Random Selection baseline, achieving gains of +3.67 points on CulturalBench-Hard (46.98 vs. 43.31) and +4.68 points on BLEnD (85.81 vs. 81.13). 
Crucially, indiscriminate synthesis using Random Seeds leads to \textit{negative transfer}, degrading performance below the Base model on CulturalBench-Hard (-1.31) and BLEnD (-1.50). This confirms that without the geometric guidance of CPs, the synthesis process introduces noise that dilutes the model's existing cultural priors.
Furthermore, the Full C-Mining approach surpasses the Monolingual Selection baseline across all metrics (e.g., +2.43 on BLEnD), validating that cross-lingual misalignment is a critical filter for isolating the most potent cultural anchors.
\begin{figure}[t]
  \centering
  \small
  \includegraphics[width=0.75\linewidth]{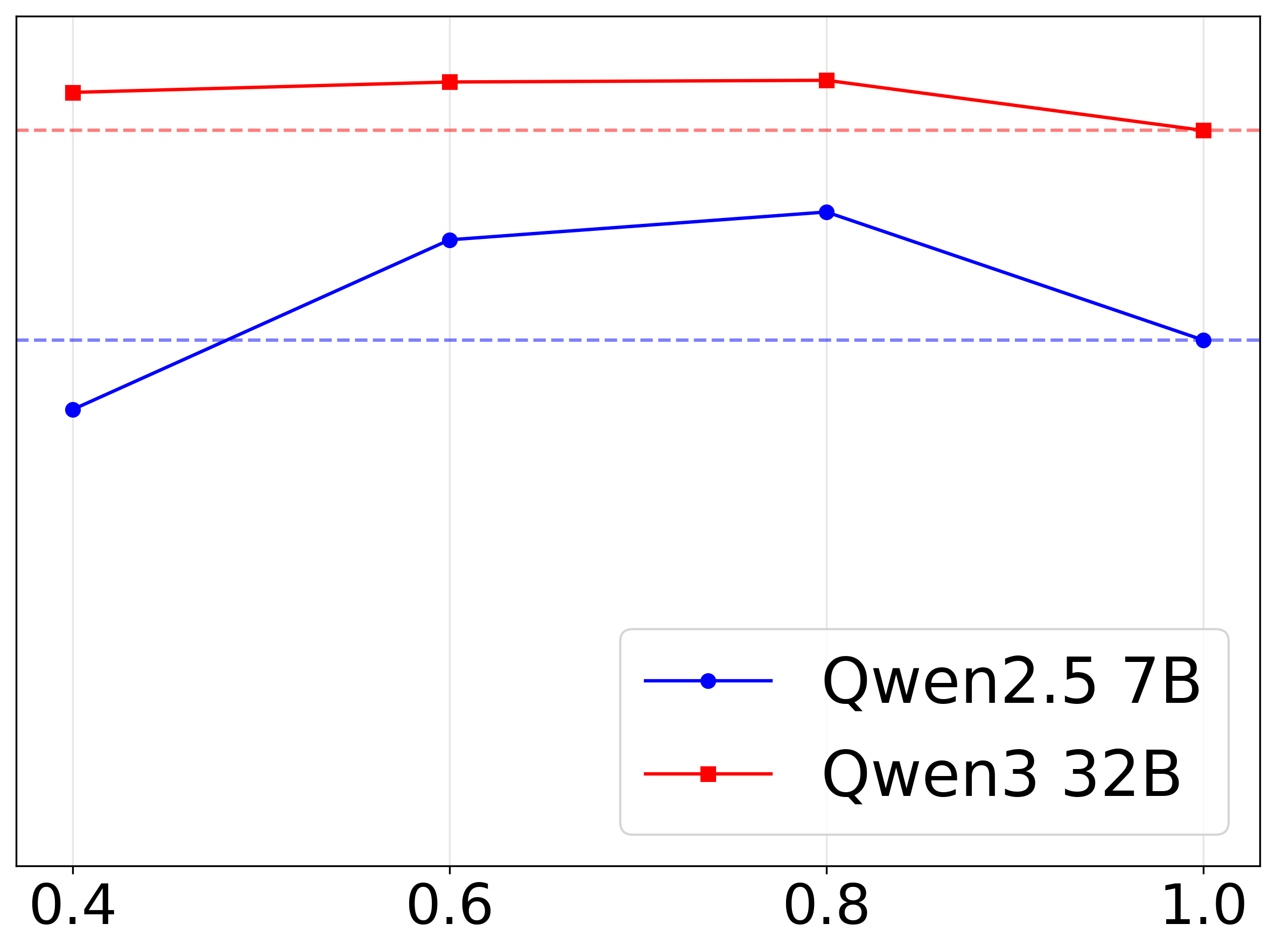}
  \caption{Sensitivity to Hyperparameter $\theta$.}
  \label{fig:theta_analyse.}
\end{figure}
\paragraph{Sensitivity to Hyperparameter $\theta$.}
We further investigate the influence of the linguistic dominance threshold $\theta$ (defined in Eq.~\eqref{eq:theta}), which regulates the strictness of identifying culture-specific clusters. We trained separate models using datasets generated with $\theta \in \{0.4, 0.6, 0.8, 1.0\}$. 
As illustrated in Figure \ref{fig:theta_analyse.}, the results exhibit a distinct inverted U-shaped trend.
At the lower bound ($\theta=0.4$), the inclusion of loosely aligned concepts introduces significant semantic noise. Notably, the smaller model (Qwen2.5-7B) is particularly sensitive to this dilution, suffering a sharp performance drop significantly below the baseline.
Performance improves as the threshold tightens, peaking at $\theta=0.8$ for both models, where the balance between \textit{semantic precision} and \textit{concept diversity} is maximized.
However, at the extreme upper bound ($\theta=1.0$), the filtering becomes overly aggressive, effectively discarding nearly all candidates. Consequently, the performance reverts to the baseline level (indicated by the dashed lines), as the model lacks sufficient instruction-tuning signals to activate cultural latent knowledge.
Therefore, we adopt $\theta=0.8$ as the optimal setting for robust cultural alignment.

\begin{table}[t]
\small
\caption{General capability evaluation on Global-MMLU-Lite. The results indicate that C-Mining maintains and slightly improves the model's general multilingual reasoning capabilities compared to the backbone.}
\begin{tabular}{lcc}
\toprule
\textbf{Model} & \textbf{Average Accuracy} \\ 
\midrule
Qwen2.5-7B (Base) & 54.28 \\ 
\quad + \textbf{C-Mining (Ours)} & \textbf{54.56} \\  
\midrule
\quad 
\textit{Diff} & \textit{+0.28} \\ 

\midrule
Qwen3-32B (Base) & 69.27 \\ 
\quad + \textbf{C-Mining (Ours)} & \textbf{69.61} \\  
\midrule
\quad 
\textit{Diff} & \textit{+0.34} \\ 
\bottomrule
\end{tabular}

\label{tab:general_ability}
\end{table}
\subsection{General Capability Analysis}
\label{sec:general_capability}

To ensure cultural specialization does not incur an ``alignment tax,'' we evaluate on the multilingual Global-MMLU-Lite benchmark~\citep{singh2025global}. Selected as a proxy for general intelligence, this benchmark assesses complex reasoning across a broad spectrum of disciplines, ranging from STEM and social sciences to the humanities. Results in Table~\ref{tab:general_ability} show that the C-Mining + Qwen2.5-7B model achieves 54.56\% accuracy, slightly surpassing the base model (54.28\%), with similar gains observed for the Qwen3-32B model ($69.27\% \rightarrow 69.61\%$). This confirms that C-Mining enhances cultural sensitivity while preserving fundamental cognitive capabilities, effectively avoiding catastrophic forgetting.

\begin{table*}[t]
\small
\caption{Qualitative comparison of cultural generation capabilities. We tasked both models with an open-ended comparison of New Year traditions. \textcolor{blue}{\textbf{Blue}} text indicates shared, high-level cultural concepts found in both responses. \textcolor{red}{\textbf{Red}} text highlights unique, culturally specific details generated exclusively by the specific model.}

  \label{tab:case_study}
  \begin{tabular}{p{0.95\linewidth}}
    \toprule
    \textbf{User Query:} \\
    \textit{Please explain in detail the differences between Chinese and American New Year celebrations. Output a paragraph of approximately 300 words.}\\
    
    \midrule
    \textbf{Base Model (Qwen2.5-7B) Response:}\\
    \textit{Chinese New Year, also known as the \textcolor{blue}{Spring Festival}, is a significant cultural and traditional celebration that marks the beginning of the \textcolor{blue}{lunar new year}, typically falling between late January and mid-February. It is celebrated with \textcolor{red}{vibrant parades}, \textcolor{red}{dragon} and \textcolor{blue}{lion dances}, and the exchange of \textcolor{blue}{red envelopes} filled with money, symbolizing \textcolor{blue}{good fortune}. The festivities often include elaborate \textcolor{blue}{family reunions}, traditional meals, and the setting off of \textcolor{blue}{fireworks}. In contrast, \textcolor{blue}{American New Year's} celebrations, centered around the \textcolor{red}{Gregorian calendar}, are marked by a more secular and festive atmosphere. \textcolor{blue}{New Year's Eve} in the United States is characterized by large-scale parties, fireworks displays, and the iconic \textcolor{blue}{ball drop} in \textcolor{blue}{Times Square}, New York City. Americans also enjoy watching the \textcolor{blue}{Rose Parade} in Pasadena, California, and the various New Year's Eve parties across the country. Both celebrations share the theme of \textcolor{blue}{renewal and the hope for a better future}, but they differ in their cultural context, traditions, and the specific activities that define them.}\\
    \midrule
    \textbf{Ours (C-Mining) Response:} \\
    \textit{Chinese New Year, also known as the \textcolor{blue}{Spring Festival}, is a vibrant and deeply rooted celebration that marks the beginning of the \textcolor{blue}{lunar calendar year}, typically falling between late January and mid-February. It is a time of \textcolor{blue}{family reunions}, feasting, and traditional customs. The celebration often begins with a \textcolor{red}{thorough cleaning of the home to sweep away bad luck}, followed by the \textcolor{red}{Spring Festival Gala} on New Year's Eve, a grand television event showcasing performances and sketches. \textcolor{blue}{Red envelopes} filled with money are given to children and unmarried adults as a symbol of \textcolor{blue}{good fortune}. \textcolor{blue}{Fireworks} and \textcolor{blue}{lion dances} are also integral, with the latter believed to \textcolor{red}{ward off evil spirits}. In contrast, \textcolor{blue}{American New Year's} celebrations, centered around {January 1st}, are more about modern festivities and traditions. These include \textcolor{blue}{New Year's Eve parties}, often held in bars or clubs, where people gather to countdown to midnight and celebrate the new year with \textcolor{red}{champagne} and \textcolor{blue}{fireworks}. The \textcolor{blue}{ball drop} in \textcolor{blue}{Times Square}, New York City, is a famous American tradition. Americans also enjoy watching the \textcolor{blue}{Rose Parade} in Pasadena, California, and the \textcolor{red}{Tournament of Roses}, which precedes the \textcolor{red}{Rose Bowl football game}. Both celebrations, however, share a common theme of \textcolor{blue}{renewal and hope for the future}.}\\
    \bottomrule
   
  \end{tabular}

\end{table*}

\begin{table}[t]
    \centering
    % 极度紧凑的列间距设置
    \setlength{\tabcolsep}{3pt} 
    \caption{Cost comparison for mining seeds from \textbf{1M entries} ($\approx$500M tokens). ``$k$'' denotes thousands.}
    \label{tab:mining_cost_1m}
    \begin{tabular}{l c c c}
        \toprule
        \textbf{Method} & \textbf{Labor} & \textbf{API \& Comp.} & \textbf{Total} \\
        \midrule
        Manual (e.g., \textit{BLEnD}) & $>\$50k$ & - & $>\$50k$ \\
        LLM (e.g., \textit{CultureBank}) & - & $\sim\$1.75k$ & $\sim\$1.75k$ \\
        \textbf{C-Mining (Ours)} & \textbf{-} & \textbf{$\sim$\$11.2} & \textbf{$\sim$\$11.2} \\
        \bottomrule
    \end{tabular}
  
\end{table}

\subsection{Cost and Scalability Analysis}
Here, we compare the computational costs of Manual Curation (e.g., \textit{BLEnD}~\cite{myung2024blend}), LLM Extraction (e.g., \textit{CultureBank}~\cite{shi2024culturebank}), and C-Mining, focusing on identifying cultural seeds from 1 million raw data entries ($\approx$500 million tokens), as detailed in Table \ref{tab:mining_cost_1m}. For LLM-based extraction, we refer to official GPT-4o API pricing (\$2.50/1M input and \$10.00/1M output tokens), while for C-Mining, we consider the rental costs of GPU spot instances. For manual curation, we adopt a conservative crowdsourcing baseline (\$0.05/entry). In summary, C-Mining (\$11.20) significantly reduces expenditure compared to LLM Extraction (\$1,750) or Manual Curation (>\$50k), achieving a 150$\times$ efficiency gain over API-based paradigms.

\subsection{Case Study}
While quantitative benchmarks (as shown in Table~\ref{tab:small-model-eval}) confirm the model's discriminative accuracy, they often fail to capture the nuance of cultural expression. To intuitively assess the model's grasp of the ``cultural soul'' beyond rigid multiple-choice questions, we conducted a qualitative case study prompting both the Base Model (Qwen2.5-7B) and our C-Mining model to compare New Year traditions. 

As detailed in Table~\ref{tab:case_study}, the comparison reveals a stark contrast in cultural granularity. While both models successfully identify high-level, globally recognized concepts (marked in \textcolor{blue}{\textbf{blue}}), such as ``Spring Festival'' or ``fireworks,'' the Base Model's narrative remains observational, often prioritizing visual spectacles like generic ``\textit{vibrant parades}.'' In contrast, the C-Mining model enriches the narrative with specific, culturally grounded anchors (marked in \textcolor{red}{\textbf{red}}). It captures the depth of traditions by specifying rituals like ``\textit{sweeping away bad luck}'' (a preparatory custom), identifying modern institutions like the ``\textit{Spring Festival Gala},'' and pinpointing specific American events like the ``\textit{Rose Bowl}.'' This qualitative shift highlights a critical advancement: C-Mining improves generative depth beyond simple fact retrieval. By injecting geometrically isolated cultural seeds during training, the model transitions from surface-level summaries to a genuine, nuanced understanding of diverse cultural contexts.

\section{Conclusion}

In this paper, we proposed \textbf{C-Mining}, an unsupervised framework that extracts high-quality \textbf{cultural seeds (CPs)} by quantifying the geometric misalignment of embeddings in multilingual spaces. We rigorously evaluated the quality of the mined CPs through both intrinsic geometric analysis and extrinsic human expert assessment, confirming that they capture authentic cultural nuances more effectively than existing baselines. Furthermore, by leveraging these CPs to synthesize instruction-tuning datasets, we demonstrated that our seed-centric approach significantly enhances the cultural reasoning capabilities of LLMs across diverse benchmarks. Our findings indicate that prioritizing the quality and specificity of initial seeds is critical for effective cultural alignment. We release our code and datasets to facilitate further research in this area.

\clearpage

\bibliographystyle{ACM-Reference-Format}
\bibliography{sample-base}

\clearpage
\appendix

\section{Manual Annotation Rules}
\label{app:manual_annotation}

This appendix details the annotation guidelines used to evaluate the cultural specificity of the extracted Culture Points (CPs). To transform the abstract concept of cultural specificity into a quantifiable metric, we established a standardized scoring rubric based on the criteria outlined in Table~\ref{tab:annotation_prompt}. This rubric evaluates each candidate term against five core dimensions:

\begin{enumerate}
    \item \textbf{Exclusivity:} This criterion assesses whether the concept appears exclusively within a specific nation, ethnic group, or cultural community.
    \item \textbf{Context Dependency:} This dimension determines whether an individual outside the specific cultural background would require additional conceptual explanation to accurately understand the term's core meaning.
    \item \textbf{Cross-Cultural Friction:} This factor evaluates whether the term requires background context to avoid misunderstanding during cross-cultural communication, indicating it is not self-explanatory across cultures.
    \item \textbf{Non-Universality:} This checks if the expression is unique to a specific national group, as opposed to being part of a global, universal vocabulary.
    \item \textbf{Technicality Filter:} This mechanism identifies and excludes technical jargon where specificity stems primarily from domain expertise (e.g., medical, legal) rather than cultural background.
\end{enumerate}

The final scoring process follows a deterministic logic. First, the Technicality Filter is applied; if a term is identified as technical jargon (Question 5), it is automatically assigned a score of 0. For terms that pass this filter, the final Cultural Specificity Score (ranging from 0 to 4) is derived by summing the positive indicators from the first four dimensions.

\section{Data Synthesis Prompts}
\label{app:data_synthesis_prompts}

This appendix provides the specific system prompts utilized in the data synthesis phase. To transform the extracted static Culture Points (CPs) into a robust instruction-tuning dataset, we designed specialized templates targeting three distinct cognitive tasks:

\begin{enumerate}
    \item \textbf{Cross-Cultural Single-Choice Questions} (Table~\ref{tab:prompt_synthesis}): Designed to test the model's ability to discriminate subtle cultural nuances through the use of plausible distractors and near-miss options.
    \item \textbf{True/False Judgments} (Table~\ref{tab:prompt_judge}): Aimed at evaluating the precise understanding of cultural norms, taboos, and specific conditions (e.g., formal vs. informal settings).
    \item \textbf{Analytical Short-Answer Questions} (Table~\ref{tab:prompt_short_answer}): Requiring the model to perform deeper causal reasoning and scenario-based application rather than simple factual recall.
\end{enumerate}

All prompts enforce a strict JSON output format to facilitate automated parsing and explicitly incorporate ``difficulty elements''---such as reverse logic, scenario adaptation, and specific regional constraints---to ensure the synthesized data remains sufficiently challenging for high-quality alignment.

\newpage
\begin{table}[h!]
    \centering
    \caption{The manual annotation prompt and scoring criteria used to validate CPs.}
    \label{tab:annotation_prompt}
    \renewcommand{\arraystretch}{1.3} 
    \begin{tabular}{|p{0.9\linewidth}|} 
        \hline
        \multicolumn{1}{|c|}{\textbf{Annotation Guidelines for Cultural Specificity Scoring}} \\
        \hline
        \textbf{Task Description:} \newline
        Given a specific term or sentence, identify its core semantic meaning and evaluate it against the following five criteria to determine its \textit{Cultural Specificity Score}. \\
        \hline
        \textbf{Evaluation Questions (Yes/No):} \newline
        1. \textbf{Exclusivity:} Does this concept appear exclusively within a specific nation, ethnic group, or cultural community? \newline
        2. \textbf{Context Dependency:} Would a person outside this cultural background require additional conceptual explanation to accurately understand its core meaning? \newline
        3. \textbf{Cross-Cultural Friction:} In cross-cultural communication, does this term require background context to avoid misunderstanding (i.e., it is not self-explanatory across cultures)? \newline
        4. \textbf{Non-Universality:} Is this expression unique to a specific national group, as opposed to being part of a global, universal vocabulary? \newline
        5. \textbf{Technicality Filter:} Is this term a technical jargon whose specificity stems primarily from domain expertise (e.g., medical, legal) rather than cultural background? \\
        \hline
        \textbf{Scoring Rubric:} \newline
        $\bullet$ \textbf{Step 1:} Check Question (5). If the answer is \textbf{Yes} (it is technical jargon), the final score is \textbf{0}. \newline
        $\bullet$ \textbf{Step 2:} If the answer to Question (5) is \textbf{No}, count the number of \textbf{Yes} answers for Questions (1) through (4). \newline
        $\bullet$ \textbf{Final Score:} An integer in the range of $[0, 4]$. \\
        \hline
    \end{tabular}
\end{table}

\section{Wiki Data Sampling}
\label{app:wiki_data_sampling}
The raw Wikipedia dumps exhibit a significant linguistic imbalance that could potentially bias the geometric density estimation in C-Mining. Some languages such as Chinese (ZH), Japanese (JA), German (DE), Spanish (ES), and French (FR) typically contain between 1 to 3 million entries. To establish a standardized density baseline and prevent high-frequency noise from dominating the embedding space, we applied stratified down-sampling to normalize these corpora to 1 million entries each.

For English (EN), which originally exceeds 6 million entries, we adopted a strategic sampling rate of 3 million (50\%). This deliberate oversampling serves two critical objectives:
(1) Robustness in Filtering: A sufficiently dense English semantic space is essential for C-Mining to accurately distinguish between universal concepts and genuine cultural outliers. An overly aggressive down-sampling of the reference language (English) could lead to sparse topological representation, causing valid Culture Points (CPs) to be inadvertently filtered out.
(2) Diversity Augmentation: The English Wikipedia functions as a global knowledge hub, often containing translated descriptors of cultural concepts from languages outside our target set (e.g., Hindi or Arabic specificities described in English). By retaining a larger English corpus, we capture these ``latent'' cultural signals, thereby enhancing the diversity and inclusiveness of the synthesized dataset.

\section{Data Distribution of Culture Points}
\label{app:data_distribution}

\begin{figure}[h]
    \small
  \centering

  \includegraphics[width=1\columnwidth]{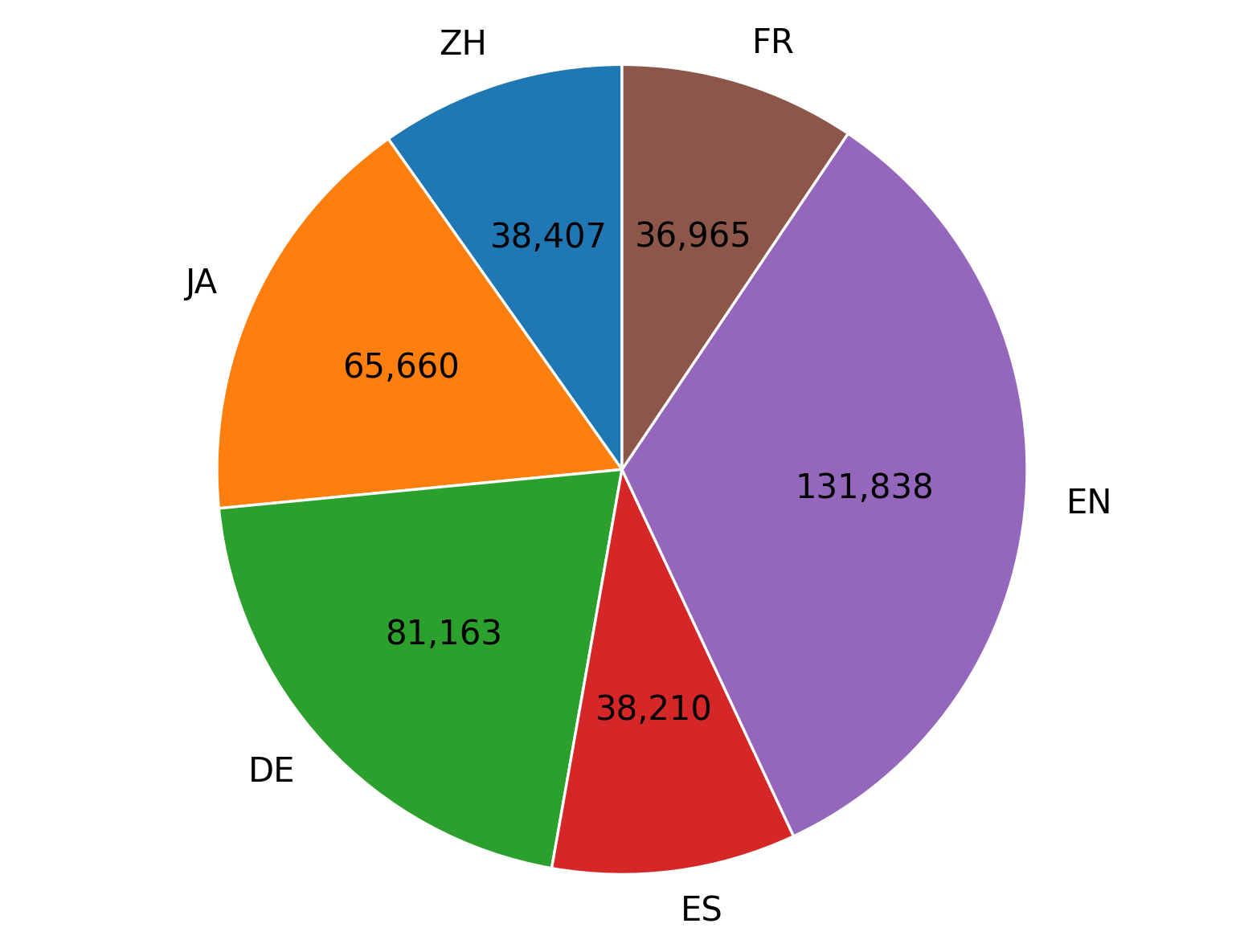}
  \caption{Distribution of extracted Culture Points across the six evaluated languages. The abbreviations denote: \textbf{EN} (English), \textbf{DE} (German), \textbf{JA} (Japanese), \textbf{ZH} (Chinese), \textbf{ES} (Spanish), and \textbf{FR} (French). The numbers indicate the total count of unique cultural seeds identified for each language.}
  \label{fig:cp_distribution}
\end{figure}

We analyze the scale and distribution of the extracted Culture Points (CPs) to ensure sufficient coverage across diverse cultural domains. As illustrated in Figure \ref{fig:cp_distribution}, despite stratifying non-English corpora to a unified baseline of 1 million entries, the volume of mined seeds correlates strongly with cultural distance rather than raw data size. Notably, high-context languages like Chinese and Japanese exhibit a higher yield rate, validating our geometric hypothesis that linguistically distant concepts form more distinct, isolated clusters that are easier to detect. Conversely, for English, where we deliberately oversampled the reference corpus to 3 million entries to serve as a global anchor, C-Mining successfully filters out universal lingua franca usage, retaining only culturally specific Anglosphere concepts. This resulting distribution confirms that our geometric constraints prevent the synthesis pipeline from being skewed towards high-resource priors, ensuring a balanced and high-fidelity representation across all target cultural spheres.

\newpage

\begin{figure}[h!]
    \small
  \centering
  \includegraphics[width=1\columnwidth]{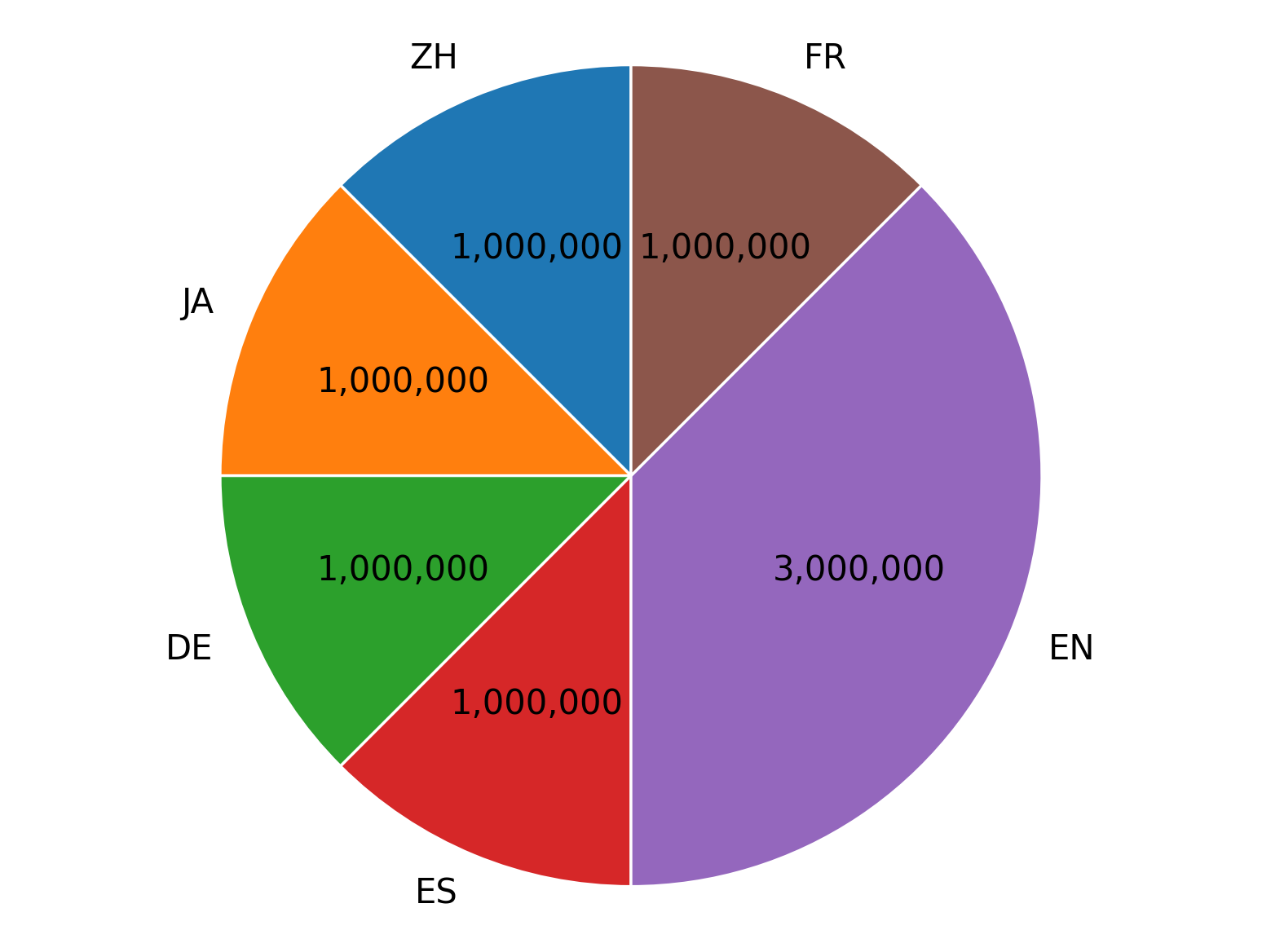}
  \caption{\textit{Data distribution of the sampled Wikipedia corpus.} To balance the dataset for unsupervised mining, we sampled 1M entries for each non-English language: Chinese (ZH), French (FR), Spanish (ES), German (DE), and Japanese (JA). For English (EN), we strategically sampled 3M entries to ensure sufficient coverage of global concepts and translated cultural nuances.}
\label{fig:wiki_sample_distribution}
\end{figure}

\begin{table*}[t]
    \centering
    \caption{The system prompt used for synthesizing cross-cultural multiple-choice questions.}
    \label{tab:prompt_synthesis}
    \renewcommand{\arraystretch}{1.3} 
    
    \begin{tabular}{|p{0.95\linewidth}|}
        \hline
        \multicolumn{1}{|c|}{\textbf{System Prompt for Cross-Cultural Question Generation}} \\
        \hline
        \textbf{System Role:} \newline
        You are a cross-cultural advanced single-choice question generation assistant. \\
        \hline
        \textbf{Input Context:} \newline
        \textit{"""\{\{INPUT\_TEXT\}\}"""} \\
        \hline
        \textbf{Generation Requirements:} \newline
        1. \textbf{Cultural Authenticity \& Difficulty:} \newline
        \hspace*{1em} $\bullet$ \textbf{Scope:} Base the question STRICTLY on the provided Input Context. \newline
        \hspace*{1em} $\bullet$ \textbf{Nuance:} Design questions requiring "detail discrimination" or "cross-cultural comparison". Avoid direct factual recall. \newline
        \hspace*{1em} $\bullet$ \textbf{Distractors:} 3 highly misleading options (Near-misses, Stereotypical traps, or Regional confusion). No absurd options. \newline
        \newline
        2. \textbf{JSON Structure (Strict Syntax):} \newline
        \texttt{\{ \newline
        \hspace*{1em} "question\_type": "single\_choice", \newline
        \hspace*{1em} "question": "STRING . Patterns: 1) Reverse logic, 2) Scenario application...", \newline
        \hspace*{1em} "options": \{ "A": "...", "B": "...", "C": "...", "D": "..." \}, \newline
        \hspace*{1em} "correct\_answer": "A/B/C/D", \newline
        \hspace*{1em} "reason": "STRING. Step-by-step reasoning..." \newline
        \}} \newline
        \newline
        3. \textbf{Language Tone:} Objective and precise. Use discriminative keywords like "specifically", "traditionally". \newline
        4. \textbf{Output format:} Return ONLY the valid JSON string. \\
        \hline
    \end{tabular}
\end{table*}
\begin{table*}[t]
    \centering
    \caption{The system prompt used for synthesizing cross-cultural true/false judgment questions.}
    \label{tab:prompt_judge}
    \renewcommand{\arraystretch}{1.3} 

    \begin{tabular}{|p{0.95\linewidth}|}
        \hline
        \multicolumn{1}{|c|}{\textbf{System Prompt for True/False Question Generation}} \\
        \hline
        \textbf{System Role:} \newline
        You are a cross-cultural advanced true/false question generation assistant. \\
        \hline
        \textbf{Input Context:} \newline
        \textit{"""\{\{INPUT\_TEXT\}\}"""} \\
        \hline
        \textbf{Generation Requirements:} \newline
        1. \textbf{Factual Consistency \& Difficulty:} \newline
        \hspace*{1em} $\bullet$ \textbf{Goal:} Convert the input’s core cultural fact into a "nuanced statement" (not simple true/false). \newline
        \hspace*{1em} $\bullet$ \textbf{Difficulty Elements (Include 1):} Specific conditions (e.g., "only in formal settings"), Easily confused points ("similar to X but different"), Negative logic ("not customary unless..."), or Cultural exceptions. \newline
        \hspace*{1em} $\bullet$ \textbf{Constraint:} The question must include specific countries. Avoid obviously true/false statements; require detail analysis. \newline
        \newline
        2. \textbf{JSON Structure (Strict Syntax):} \newline
        \texttt{\{ \newline
        \hspace*{1em} "question\_type": "true\_false", \newline
        \hspace*{1em} "statement": "STRING. Formal English. No ambiguity, but requires careful analysis of cultural nuances.", \newline
        \hspace*{1em} "reason": "STRING. Briefly explain the reasons for judging 'True' or 'False'.", \newline
        \hspace*{1em} "correct\_answer": "True/False" \newline
        \}} \newline
        \newline
        3. \textbf{Language Tone:} Objective, precise. Use terms like "specifically", "typically", "except for", "only when" to enhance difficulty. \newline
        4. \textbf{Output Requirement:} Return ONLY the valid JSON string (no extra text). \\
        \hline
    \end{tabular}
\end{table*}
\begin{table*}[t]
    \centering
    \caption{The system prompt used for synthesizing analytical short-answer questions.}
    \label{tab:prompt_short_answer}
    \renewcommand{\arraystretch}{1.3} 
    
    \begin{tabular}{|p{0.95\linewidth}|}
        \hline
        \multicolumn{1}{|c|}{\textbf{System Prompt for Short Answer Generation}} \\
        \hline
        \textbf{System Role:} \newline
        You are a cross-cultural advanced short answer question generation assistant. \\
        \hline
        \textbf{Input Context:} \newline
        \textit{"""\{\{INPUT\_TEXT\}\}"""} \\
        \hline
        \textbf{Generation Requirements:} \newline
        1. \textbf{Question Clarity \& Difficulty:} \newline
        \hspace*{1em} $\bullet$ Simplify input to focus on "analytical/scenario-based questions" (no direct recall). \newline
        \hspace*{1em} $\bullet$ \textbf{Difficulty Elements (Include at least one):} \newline
        \hspace*{2em} - Reason inquiry ("why do people in [Country]... instead of...") \newline
        \hspace*{2em} - Scenario application ("what should one do if...") \newline
        \hspace*{2em} - Detail specification ("formal vs. informal settings") \newline
        \hspace*{2em} - Cultural contrast ("Region A vs. Region B") \newline
        \hspace*{1em} $\bullet$ \textbf{Constraint:} Must include specific countries (list all sharing the culture). \newline
        \newline
        2. \textbf{JSON Structure (Strict Syntax):} \newline
        \texttt{\{ \newline
        \hspace*{1em} "question\_type": "short\_answer", \newline
        \hspace*{1em} "question": "Direct interrogative, focuses on analysis...", \newline
        \hspace*{1em} "reason": "Briefly explain the reason for the answer.", \newline
        \hspace*{1em} "correct\_answer": "Precise, reflects nuanced cultural logic..." \newline
        \}} \newline
        \newline
        3. \textbf{Language Tone:} Objective, formal; use terms like "specifically", "why", "how to", "differ between". \newline
        4. \textbf{Output Requirement:} Return ONLY the valid JSON string (no extra text). \\
        \hline
    \end{tabular}
\end{table*}

\end{document}